\documentclass[letterpaper]{article}
\usepackage{proceed2e}
\usepackage[margin=1in]{geometry}
\usepackage[round]{natbib} 
\usepackage[linesnumbered , ruled, noend, norelsize]{algorithm2e}
  \SetKwInOut{Input}{Input}
  \SetKwInOut{Output}{Output}
\IncMargin{1em}

\usepackage{amssymb, verbatim, bm, amsmath, float, graphicx}
\allowdisplaybreaks

\newcommand{\calT}{\mathcal{T}}
\newcommand{\calC}{\mathcal{C}}
\newcommand{\calP}{\mathcal{P}}
\newcommand{\calD}{\mathcal{D}}

\usepackage{times}

\title{Propagation for Dynamic Continuous Time Chain Event Graphs}

\author{ {\bf Aditi Shenvi} \\
Centre for Complexity Science, \\
University of Warwick\\
\& The Alan Turing Institute.\\
\And
{\bf Jim Q. Smith}  \\
Department of Statistics, \\
University of Warwick\\
\& The Alan Turing Institute.\\
}

\begin{document}

\maketitle

\begin{abstract}
Chain Event Graphs (CEGs) are a family of event-based graphical models that represent context-specific conditional independences typically exhibited by asymmetric state space problems. The class of continuous time dynamic CEGs (CT-DCEGs) provides a factored representation of longitudinally evolving trajectories of a process in continuous time. Temporal evidence in a CT-DCEG introduces dependence between its transition and holding time distributions. We present a tractable exact inferential scheme analogous to the scheme in \citet{kjaerulff1992computational} for discrete Dynamic Bayesian Networks (DBNs) which employs standard junction tree inference by ``unrolling" the DBN. To enable this scheme, we present an extension of the standard CEG propagation algorithm \citep{thwaites2008propagation}. Interestingly, the CT-DCEG benefits from simplification of its graph on observing compatible evidence while preserving the still relevant symmetries within the asymmetric network. Our results indicate that the CT-DCEG is preferred to DBNs and continuous time BNs under contexts involving significant asymmetry and a natural total ordering of the process evolution.   
\end{abstract}
%This is particularly of interest when the time evolution of the network is dependent on the edges traversed by the unit which is the case in many real-world scenarios. 

\section{INTRODUCTION}
%few lines of why the CEG and event tree are used as against BNs/other models.
Real-world systems often contain asymmetric state spaces and do not admit a natural product structure. To model such systems with traditional graphical models such as Bayesian Networks (BNs), including its various static and dynamic variants, a set of distinct measurement variables first need to be elicited. For many asymmetric processes, this is not always easy, natural or defensible to do. Forcing a random variable based description of such systems results in a model that stores lots of redundant structural zeroes within its conditional probability tables (CPTs). For instance, a variable describing the post-operative health condition of a patient does not make sense if the patient died during the operation.

Additionally, asymmetric systems often exhibit context-specific conditional independences where the independence relationship depends on the realisations of the conditioning set. Modifications of the BN have been proposed for embodying context-specific independences (for e.g., \citet{geiger1996knowledge, boutilier1996context, poole2003exploiting}); often using tree-like adaptations of the graph or the CPTs in the process. 

Chain Event Graphs (CEGs) \citep{collazo2018chain} address both these shortcomings of BNs when it comes to asymmetric processes. CEGs are built from event trees which provide an excellent and natural framework to describe the evolution of such processes \citep{shafer1996art}. Through a series of transformations involving the colouring of its vertices, an event tree is transformed into the more concise graph of a CEG. The topology of the CEG describes the partial or complete symmetries within the different ways in which a process might evolve. This allows reading of conditional independences, causal exploration and moreover, the graphical properties of the CEG can be drawn directly from natural language descriptions provided by domain experts before the tree is populated with probabilities.

\begin{figure}[ht]
\vspace{1in}
\centering
\includegraphics[trim = 8cm 8cm 8cm 8cm, scale = 0.37 ]{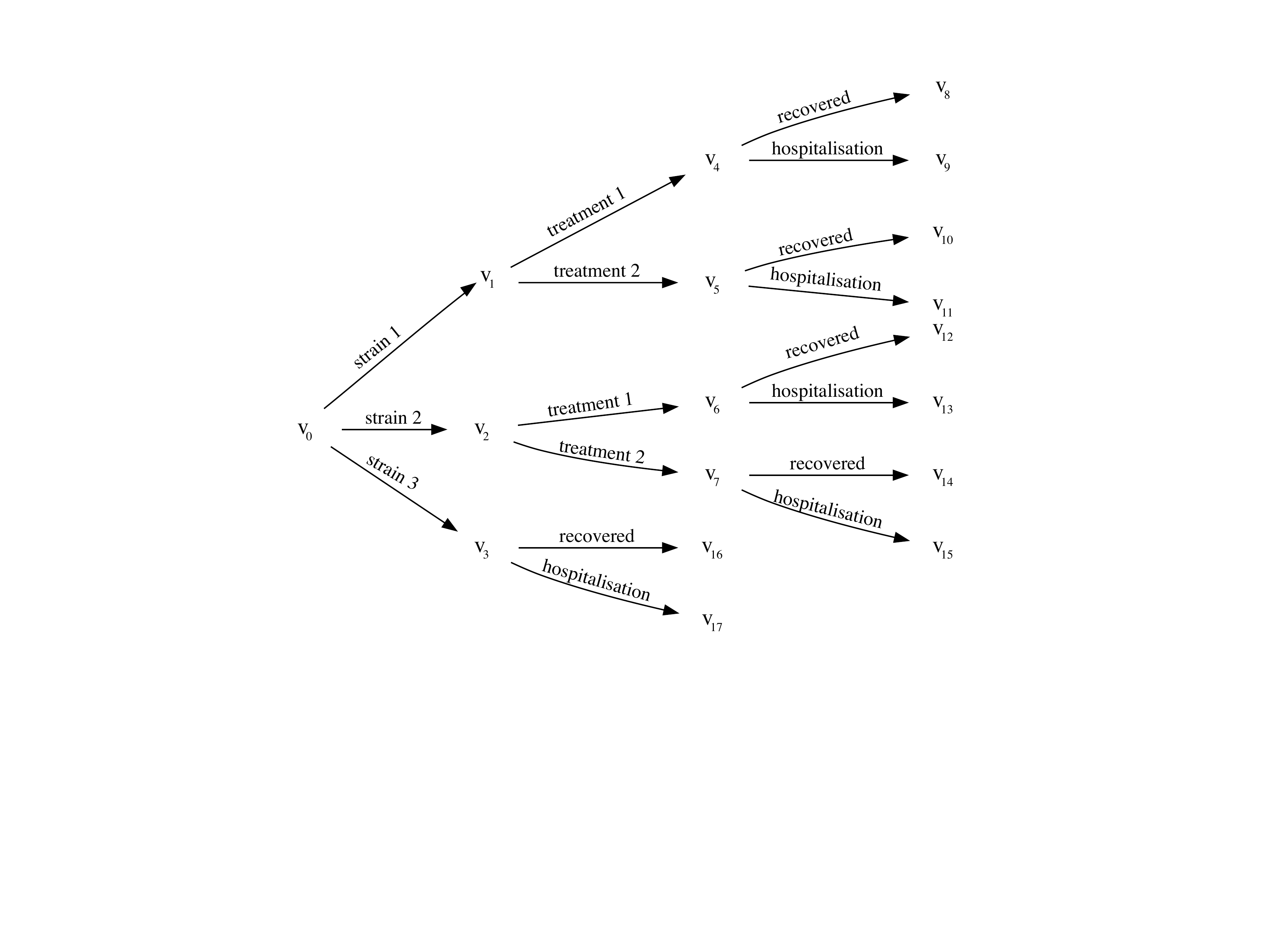}
\caption{Event tree for Example 1 and fragment of the event tree in Example 2}
\label{fig:event_tree}
\end{figure}

\textbf{Example 1:} The hypothetical event tree in Figure \ref{fig:event_tree} shows the different infection strains, treatment alternatives and outcomes for an individual (at vertex $v_0$) who shows the first symptoms of a particular infection.

%what are DCEGs? types?
The CEG family contains several dynamic variants called DCEGs \citep{barclay2015dynamic, collazo2018n, shenvi2019bayesian}. In this paper, we look at the continuous time DCEG (CT-DCEG) which is inspired by the flexibility offered by semi-Markov processes (SMPs) in modelling non-exponentially distributed holding times at the various states. First introduced as the extended DCEG in \citet{barclay2015dynamic}, this class was developed through a special case called the reduced DCEG which was applied to modelling public health interventions \citep{shenvi2019bayesian} and criminal investigations \citep{bunnin2019bayesian}. 
%This class of models benefits from the expressiveness of CEGs and the holding time flexibility offered by SMPs.

%talk about the alternative models
CT-DCEGs differ from dynamic BNs (DBNs) \citep{nicholson1994dynamic} for the reasons stated above and additionally because the latter models discrete time processes. A continuous time analogue of BNs is given by continuous time BNs (CTBNs) \citep{nodelman2002continuous}. CTBNs were inspired by Markov processes and represent the dynamics of structured multi-component processes. The holding times at the vertices are restricted to exponential distributions (except in \citet{nodelman2005expectationmax}). The major difference between CT-DCEGs and CTBNs is that the former describes the possible life-histories of a unit within a temporal process while the latter describes a temporal process through the evolution of the components describing the process. Hence CTBNs may have variables co-evolving while in a CT-DCEG, a unit can only be in one state at a time and models the time evolution from that state. Secondly, exact inference in CTBNs is exponential in its number of components and the approximate inference techniques are either (a) variational \citep{nodelman2005expectationprop, saria2007reasoning, cohn2009mean}, typically using expectation propagation; or (b) stochastic approximations \citep{fan2008sampling, el2008gibbs, fan2010importance}; in both cases using approximations that exploit the exponential holding time constraint of such models. And lastly, under plausible assumptions, the transition probabilities and holding times can be modelled independently in a CT-DCEG.

%From the above arguments, it is easy to see that a CT-CEG differs from a dynamic BN (DBN) \citep{nicholson1994dynamic} which is a time-sliced, connected set of BNs representing a dynamic discrete-time stochastic process. It is more pertinent to compare CT-CEGs to continuous time BNs (CTBNs) \citep{nodelman2002continuous} that were introduced to represent the dynamics of structured multi-component processes evolving in continuous time. CTBNs were inspired by Markov processes and have an exponential holding time at each vertex (with the exception of \citet{nodelman2005expectationmax}). The transition of a variable from one state to the next, given the realisations of its parents is represented through a conditional intensity matrix. 

Interestingly, a visionary but currently underdeveloped class of graphical models called the Temporal Nodes BN (TNBN) \citep{arroyo1999temporal} have several similarities with the CT-DCEG. Although some properties of propagation in a TNBN are similar in intuition to that in a CT-DCEG, they are presented in a non-technical way with no formal justification. Besides, point temporal evidence cannot be propagated through the TNBN \citep{galan2002networks}. Another important difference between TNBNs and CT-DCEGs is that the former, as it uses BN semantics, has to assume that each effect is caused by only one of its parents. The CT-DCEG circumvents this problem through its event tree construction and colourings of its vertices. 

In Section \ref{sec:ct-dceg} we review the terminology and framework of the CT-DCEG class. The new inference scheme for CT-DCEGs proposed in this paper showed a gap in the existing literature which we filled in Section \ref{sec:ctceg} by presenting for the first time a static continuous time CEG (CT-CEG) and proved an extension of the standard propagation algorithm \citep{thwaites2008propagation} for this class. Our inference scheme for propagating evidence in CT-DCEGs, inspired by the scheme in \citet{kjaerulff1992computational} for DBNs, is presented in Section \ref{sec:propagation}. This splits the CT-DCEG into three sub-models; two of which have linear time complexity for propagation. While here we apply this to a continuous time setting, this scheme works exactly the same for DCEGs with discrete time domains. Later in Section \ref{sec:applications} we explore what we believe to be a highly applicable novel class of models called the \textit{mixed CEGs} whose vertices are partitioned into a set for which holding times are meaningful and the other where they are not. We briefly discuss how the methods developed in this paper extend to this new class.

\section{THE CONTINUOUS TIME DCEG}
\label{sec:ct-dceg}

\subsection{FRAMEWORK AND TERMINOLOGY}

A CT-DCEG consists of a graph describing the possible developments of a process. It exploits the symmetries within these evolutions and so provides a concise representation. It also facilitates the exploration of the underlying dependences and associates a well-defined probability model to the graph. Similar to a CEG, a CT-DCEG is constructed from an event tree, albeit one which has a continuous time domain. For clarity we have chosen to use notation as close as possible to \citet{thwaites2008propagation}.

An event tree $\calT$ is a directed graph representing the evolution of a process in continuous time with a (possibly infinite) vertex set $V(\calT)$ and edge set $E(\calT)$. The set of non-leaf vertices - called situations - is represented by $S(\calT) \subset V(\calT)$. Without loss of generality, we assume that the time spent by a unit in a situation $v_i$ is dependent on the situation it visits next, say $v_i'$. Hence, this conditional \textit{holding time} can be associated with edge $e(v_i, v_i')$ which goes from $v_i$ to $v_i'$. We denote this by variable $H(e(v_i,v_i'))$. The variable $H_{v_i}$ indicates the unconditional holding time in situation $v_i$. 

Symmetries within the event tree are expressed through vertex and edge colourings associated with partitions called stages and clusters. Two situations are in the same \textit{stage} when they can be hypothesised to share the same conditional transition distribution. For the purposes of this paper we also require that for situations in the same stage, the edges with the same estimated probability share the same edge label. However, this is not essential. In fact, the edges may be retrospectively labeled to explain the symmetries observed - one of the key features of this family of models. Similarly, two edges are in the same \textit{cluster} when they share the same holding time distribution. The vertices and edges of the event tree are coloured according to their stage and cluster memberships respectively. Such a coloured tree is referred to as a \textit{hued tree}. So the stages indicate equivalences in \textit{where} a unit goes and clusters indicate equivalences about the \textit{speed} of these transitions. Table \ref{table:stages_and_clusters} gives the non-trivial stages and clusters for the finite event tree in Example 1.

An additional partition called positions defines the vertex set of the resultant CT-DCEG. Two situations in the hued tree are in the same \textit{position} if and only if the subtrees rooted at these situations are isomorphic to each other where the isomorphism preserves structure and colouring. We denote the set of positions in $\calT$ by $W(\calT)$. 

\begin{table}[ht]
\caption{Stages and clusters in Example 1}
\label{table:stages_and_clusters}
\begin{center}
\begin{tabular}{l|l}
\multicolumn{1}{c|}{\bf Stages}  &\multicolumn{1}{c}{\bf Clusters} \\
\hline
$u_1 = \{v_1, v_2\}$         & $c_1 = \{e(v_1, v_4), e(v_2, v_6)\}$ \\
$u_2 = \{v_4, v_6\}$         & $c_2 = \{e(v_1, v_5), e(v_2, v_7)\}$ \\
$u_3 = \{v_5, v_7\}$         & $c_3 = \{e(v_4, v_8), e(v_6, v_{12})\}$ \\
                             & $c_4 = \{e(v_4, v_9), e(v_6, v_{13})\}$ \\
                             & $c_5 = \{e(v_5, v_{10}), e(v_7, v_{14})\}$ \\
                             & $c_6 = \{e(v_5, v_{11}), e(v_7, v_{15})\}$ \\
\end{tabular}
\end{center}
\end{table}

From a hued tree representation, we obtain a CT-DCEG $\calD = (V(\calD), E(\calD))$ by coalescing situations in the same position into a single vertex and by collapsing all its leaves into a single sink vertex $w_\infty$. Hence $V(\calD) = W(\calT) \cup \{w_\infty\}$. Note that $W(\calD) = W(\calT)$. Only the nodes which are in the same stage but not the same position retain their colouring in the CT-DCEG. Here we focus on CT-DCEGs with a finite representation. However, this is not necessary for our proposed scheme as the graph is first ``unrolled" (see Section \ref{subsec:unroll}) to propagate evidence. This gives our scheme the flexibility of adapting easily if and when the structure of the graph needs to be partially or fully changed at a later time.  

\begin{figure}[ht]
\vspace{1in}
\centering
\includegraphics[trim = 8cm 11cm 8cm 12cm, scale = 0.24 ]{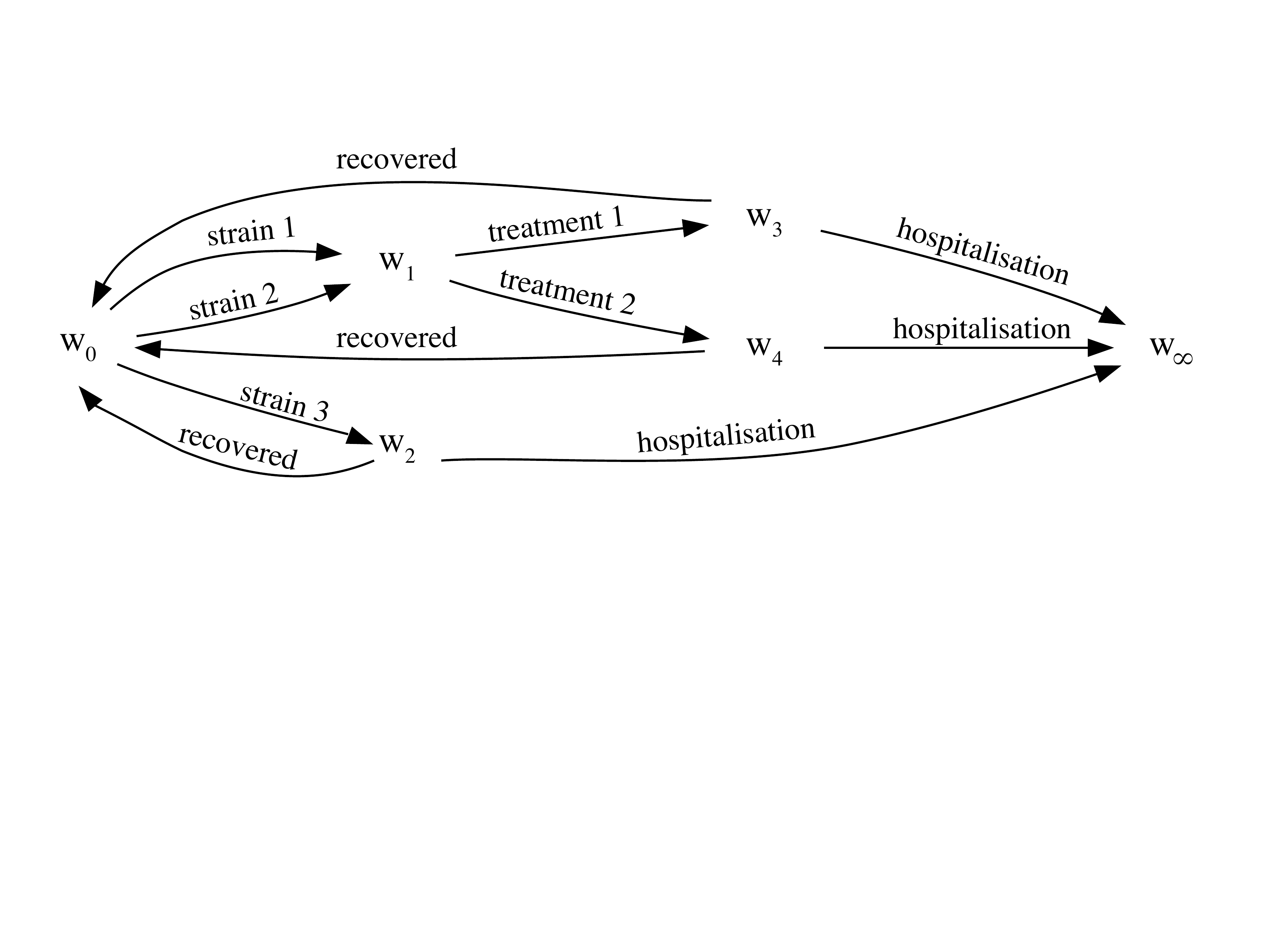}
\caption{CT-DCEG for Example 2}
\label{fig:ct-dceg}
\end{figure}

\textbf{Example 2:} Suppose we consider reinfection from the same or different strain for individuals who were not hospitalised. Assume that recovery from a particular strain does not offer any added or reduced resistance to that or the other strains. This gives an infinite event tree whose fragments are represented in Figure \ref{fig:event_tree}. The repetition in structure and probabilities results in the CT-DCEG in Figure \ref{fig:ct-dceg} where returns are represented by backward arrows labelled ``recovered".

These backward edges representing a repetition of structure in the underlying hued tree are called \textit{cyclic edges}. Additionally, we define \textit{passage-slices} which will play the role of time-slices in this continuous time setting. The first passage-slice $P_1$ is a subgraph of the CT-DCEG starting at its root and following all possible developments of a unit until it either arrives at $w_\infty$ or up to the vertex from which it traverses along a cyclic edge (e.g. $w_i, i = 2,3,4$ in Figure \ref{sec:ct-dceg}). The subsequent passage slices $P_k$ are a collection of subgraphs of the CT-DCEG such that each subgraph is rooted at a vertex into which a cyclic edge from the preceding passage-slice $P_{k-1}$ enters, $k = 2,3,\ldots$. The termination of each subgraph is determined as described for $P_1$ above. Thus, the cyclic edges connect the passage-slices. In practice, the time-interval of a passage-slice can be arbitrarily defined. In our example, all the passage-slices are identical.

The event tree notation introduced earlier extends in an obvious way to CT-DCEGs. Transition probabilities in a CT-DCEG $\calD$ can be written as the probability of an event defined using the set of its root-to-sink paths, $\calD_\Lambda$. The probability of reaching a position $w \in W(\calD)$ is given by $\pi(w) = \pi(\Lambda(w))$ where $\Lambda(w)$ is the union of all paths in $\calD_\Lambda$ passing through $w$. Similarly, the probability of passing through $e(w, w')$ is given by $\pi_e(w'\,|\, w) = \pi(\Lambda(e(w, w'))\,|\, \Lambda(w))$ where $\Lambda(e(w, w'))$ is the union of all paths in $\calD_\Lambda$ passing through the edge $e(w, w')$. The holding time density of staying at position $w$ for time $t$ before transitioning along edge $e(w, w')$ is denoted by $\pi_e^t(w' \,|\, w) = \pi(H_w = t \,|\, \Lambda(e(w, w')), \Lambda(w))$. Let $\boldsymbol{\pi}(w)= \{\pi_e(w'\,|\,w) \,|\, e(w, w') \in E(\calD)\}$ and $\boldsymbol{\pi}^t(w) = \{\pi_e^t(w'\,|\,w) \,|\, e(w, w') \in E(\calD)\}$.

\subsection{UNROLLING A CT-DCEG}
\label{subsec:unroll}

Denote by  $\epsilon_k \subset E(\calD)$ the set of cyclic edges connecting passage-slice $P_k$ to $P_{k+1}$, $k \in \mathbb{N}$. Any DCEG can be ``unrolled" into an infinitely large CEG \citep{collazo2018n, shenvi2019bayesian} by connecting its passage-slices with the corresponding cyclic edges. This is analogous to unrolling a DBN. 

\begin{figure}[ht]
\vspace{1in}
\centering
\includegraphics[trim = 8cm 16cm 8cm 13.2cm, scale = 0.37 ]{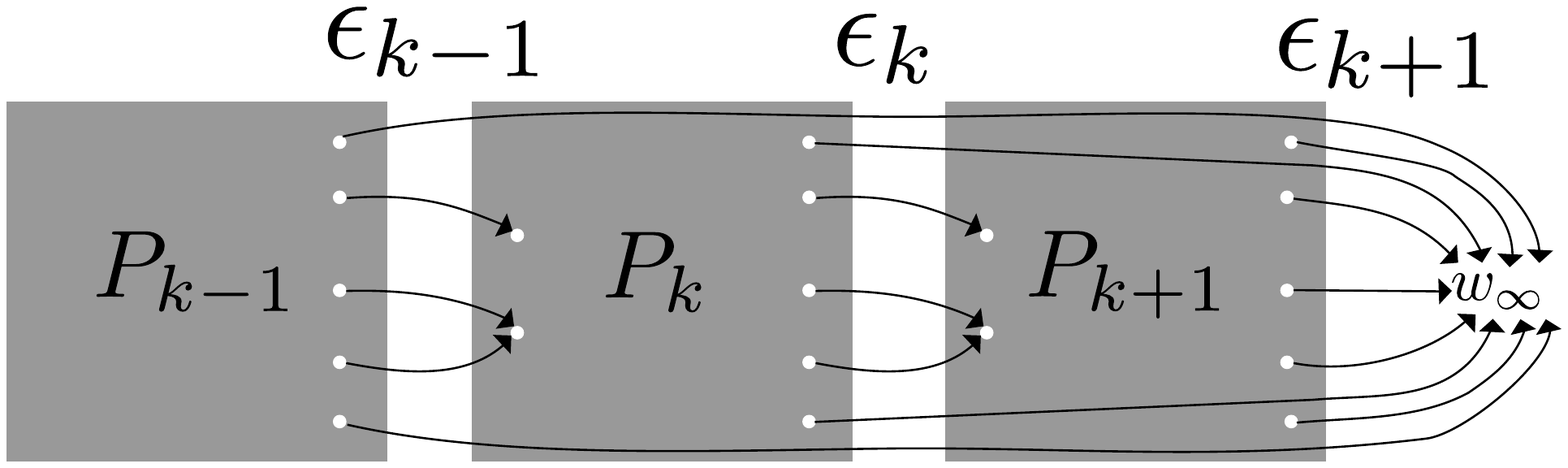}
\caption{Unrolling a CT-DCEG}
\label{fig:unrolling}
\end{figure}

Denote by $\calP_{k:k+l}$ a CT-DCEG $\calD$ which has been unrolled from passage-slices $k$ to $k+l$, for $k, l \in \mathbb{N}$. All the edges in $\epsilon_{k+l}$ are collected into a sink node $w_\infty$. The unrolling process may result in multiple sink nodes then merged into a single sink node. Figure \ref{fig:unrolling} shows a pictorial description of this process.
% Since the proposed propagation scheme is performed on the unrolled CT-DCEG, the passage-slices need not be identical. 

\subsection{SEMI-MARKOV REPRESENTATION}
\label{subsec:semi-markov}

A CT-DCEG can be represented as an SMP (see Section 5 and the relevant appendices of \citet{shenvi2019bayesian}). Informally, an SMP is a stochastic process where the next state $x_i'$ occupied by a unit in state $x_i$ is determined by the transition probabilities of its embedded Markov chain, and the distribution governing the time spent in state $x_i$ is determined by the choice of $x_i'$. The positions of a CT-DCEG can be regarded as states in its corresponding SMP. Furthermore, depending on the evidence, only a small subset of nodes of the original CT-DCEG might be relevant. Such a CT-DCEG can be represented by a condensed SMP. The state-transition diagram of the SMP for the CT-DCEG in Example 2 is identical to Figure \ref{sec:ct-dceg} with the exception that the two edges from position $w_0$ to $w_1$ are merged into a single edge with a holding time distribution that is a mixture of their individual holding time distributions. 

\section{THE CONTINUOUS TIME CEG}
\label{sec:ctceg}

A CT-CEG is a static variant of the CT-DCEG or equivalently, a continuous time analogue of the discrete time CEG. A CT-CEG is an acyclic event-based graphical model with a total ordering (coming out of its event tree construction) and vertices evolving at possibly varying time granularities - a semi-Markovian approach. It has one sink node $w_\infty$ to collect its leaves. For simplicity, here we consider time-homogeneous CT-CEGs. 

Say that a CT-CEG $\calC$ is \textit{completely specified} when
\begin{align*}
\boldsymbol{\pi} &= \{\boldsymbol{\pi}(w) \,|\, w \in W(\calC)\}, \\
\boldsymbol{\pi}^t &= \{\boldsymbol{\pi}^t(w) \,|\, w \in W(\calC), t \in \mathbb{R}^+_0\}
\end{align*}
\noindent are specified. The joint distribution of any events measurable with respect to the $\sigma$-algebra of $\calC$ can be obtained when $\calC$ is completely specified. Let $\lambda \in \calC_\lambda$ be a path of a sequence of $n(\lambda)$ edges, and $\lambda(t)$ be a sequence of times at which transitions are made along the edges of $\lambda$. We can write this as a sequence of triples of vertex, edge and time spent at the vertex before going along the edge - for example, $((w_0 = w_\lambda[0], e_\lambda[0], t_\lambda[0]), \ldots, (w_\lambda[n(\lambda)-1], e_\lambda[n(\lambda)-1], t_\lambda[n(\lambda)-1]))$. The joint probability of $\lambda$ and $\lambda(t)$ can be specified as

\begin{align*}
    \pi(\lambda, \lambda(t)) = \textstyle \prod_{i = 0}^{n(\lambda)-1} & \Big \{ \pi_{e_\lambda[i]}(w_\lambda[i+1]\,|\,w_\lambda[i]) \times \\
    & \pi_{e_\lambda[i]}^{t_\lambda[i]}(w_\lambda[i+1]\,|\,w_\lambda[i]) \Big \}.
\end{align*}

\subsection{COMPATIBLE EVIDENCE AND EVENTS}
\label{subsec:compatibility}

Evidence in a BN is typically in the form of instantiations of a subset of its variables. In a CEG, evidence takes the form of intrinsic events occurring \citep{thwaites2008propagation}. Consider an event given by $E \subseteq \calC_\Lambda$ in a CT-CEG $\calC$. Call $E$ an \textit{intrinsic event} if the subgraph of $\calC$, say $\calC^*$ induced by the root-to-sink paths of $E$ are exactly the same as the set of root-to-sink paths contained in $\calC^*_\Lambda$. With respect to temporal evidence, here we consider only point evidence. Call the temporal evidence $E$ \textit{temporally compatible} if we know all the transition times for the unit starting from the root of $\calC$ up to a certain depth. If evidence $E$ defines an intrinsic event and is temporally compatible, call it \textit{compatible evidence}. In Section \ref{subsubsec:path_probs}, we consider temporal evidence where we only know the transition time for some specific, non-root vertex or vertices. 

\subsection{A PROPAGATION ALGORITHM}
\label{subsec:algorithm}

For a process described by a CT-CEG $\calC$, compatible evidence $E$ about the temporal evolution of a unit makes the retrospective transition probabilities dependent on the corresponding holding time densities. Assume that the temporal information in $E$ gives the transition times at all vertices from the root to the sink of the CT-CEG. While we may not know the exact vertices visited by the unit, we know the time the unit made its $k$th transition. Observing $E$ typically revises the probability of not visiting several of the root-to-sink paths, either partially or entirely, to one. These paths or parts thereof can be deleted from the graph of $\calC$ to obtain a condensed representation given by the adapted CT-CEG subgraph $\calC^*$ called the \textit{transporter CT-CEG}. Note that $\calC^*_\Lambda = E_\Lambda$ where $E_\Lambda$ represents the set of paths implied by the compatible evidence $E$. The original staging structure within $\calC$ may be destroyed by the deletion of vertices and edges to obtain $\calC^*$. In this sense, $\calC^*$ only preserves the conditional independences which are still valid after observing $E$. A CT-CEG $\calC^*$ is called \textit{minimal} when it contains no two vertices in $V(\calC^*)$ that have isomorphic subtrees preserving structure and colourings. While this is not essential, we will assume from now that our transporter CT-CEG $\calC^*$ is minimal.

We describe below a two-pass backward-forward message-passing algorithm which has two main steps: a backward step to calculate the potentials and emphases, and a forward step which revises the transition probabilities. Note that $\pi(.)$ refers to probabilities in $\calC$ and $\hat{\pi}(.)$ to the updated probabilities in $\calC^*$.

Denote by $E(w)$ all the edges emanating from vertex $w \in W(\calC)$. Let $W(-1) = \{w \in W(\calC) \,|\, \forall e(w, w') \in E(w), w' = w_\infty\}$. The algorithm proceeds as follows.
\begin{enumerate}
    \item For each edge $e(w, w_\infty) \in E(w)$, $w \in W(-1)$, set the \textit{t-potential} $\tau_e(w_\infty \,|\, w)$ and \textit{h-potential} $\tau_e^{t_w}(w_\infty \,|\, w)$ as 
    \begin{align*}
       \tau_e(w_\infty \,|\, w) &= \pi_e(w_\infty \,|\, w), \\
        \tau_e^{t_w}(w_\infty \,|\, w) &= \pi_e^{t_w}(w_\infty\,|\, w),
    \end{align*}
    \noindent if $e(w, w_\infty) \in E(\calC^*)$ and zero otherwise. The holding time at $w$ is indicated by $t_w$. Set the \textit{t-emphasis} $\Phi(w)$ and \textit{h-emphasis} $\Phi^{t_w}(w)$ as follows
    \begin{align*}
        \Phi(w) &= \textstyle \sum_{e \in E(w)} \tau_e(w_\infty \,|\, w), \\
        \Phi^{t_w}(w) &= \textstyle \sum_{e \in E(w)} \tau_e(w_\infty \,|\, w)\,\tau_e^{t_w}(w_\infty \,|\, w).
    \end{align*}
    \noindent Now we say that the sink $w_\infty$ and all the positions in $W(-1)$ are \textit{accommodated}.
    
     \item For an edge $e(w, w') \in E(w)$, $w \in W(\calC)$ such that all of $w$'s children are accommodated, set the t-potential and h-potential as
    \begin{align*}
        &\tau_e(w' \,|\, w) = \pi_e(w' \,|\, w) \, \Phi(w'), \\
        &\tau_e^{t_w}(w' \,|\, w) =  \pi_e^{t_w}(w'\,|\, w),
    \end{align*}
    \noindent if $e(w, w') \in E(\calC^*)$ and zero otherwise. Set the emphases as
    \begin{align*}
        \Phi(w) &= \textstyle \sum_{e \in E(w)} \tau_e(w' \,|\, w), \\
        \Phi^{t_w}(w) &= \textstyle \sum_{e \in E(w)} \tau_e(w' \,|\, w)\,\tau_e^{t_w}(w' \,|\, w). 
    \end{align*}
    Position $w$ is accommodated when the potentials and emphases are calculated for all $e \in E(w)$.
    
    \item For all $w \in W(\calC)$, the revised conditional transition probabilities are given by
    \begin{align*}
        \hat{\pi}_e(w' \,|\, w) =
        \begin{cases}
        \dfrac{\tau_e(w' \,|\, w) \, \tau_e^{t_w}(w' \,|\, w)}{\Phi^{t_w}(w)}, & \\ \quad \quad \quad \quad \textmd{if}\, e(w, w') \in E(\calC^*)\\
        0, & \\
        \quad \quad\quad \quad \textmd{if}\, e(w, w') \notin E(\calC^*).
        \end{cases}
    \end{align*}
    \noindent Note that for the edges in the transporter $\calC^*$, the holding time densities are invariant under the compatible evidence and are simply imported from the relevant edges in $\calC$.
\end{enumerate}

A proof of this result is presented in the appendix. Let $W^{-1}(w_i)$ denote the vertices that have edges terminating in $w_i \in W(\calC)$ and $E^{-1}(w_i)$ denote the edges terminating in $w_i$. The pseudo-code for the above algorithm is given in Algorithm \ref{prop_alg_pseudo}. Here, the possible arrival time at position $w_i$ is denoted by $t_i$. Note that this algorithm is also applicable to the discrete time setting where the holding time densities are replaced by the corresponding probability mass functions.

\begin{algorithm}[ht]
    \SetAlgoLined
    \Input{$\calC$, $\boldsymbol{\pi}$, $\boldsymbol{\pi}^t$, $E$, $E_\Lambda$.}
    \Output{A completely specified CT-CEG $\calC^*$.}
    Set $A \leftarrow \emptyset$, $B \leftarrow \{w_\infty\}$, $\Phi(w_\infty) \leftarrow  1$.\\
    Initialise vectors $\boldsymbol{\hat{\pi}}$, $\boldsymbol{\hat{\pi}}^t$ of length \#$W(\calC)$.\\
    \While{$B \neq \{w_0\}$}{
    \For{$w_j \in B$}{
    \For{$w_i \in W^{-1}(w_j)$}{
    \For {$e \in E(w_i) \cap E^{-1}(w_j)$}{
    \If{$e \in E_\Lambda$}{
    $\tau_e \leftarrow \pi_e \cdot \Phi(w_j)$, $\tau^{t_i}_e \leftarrow \pi^{t_i}_e$}
    \Else{$\tau_e \leftarrow 0$, $\tau^{t_i}_e \leftarrow 0$}
    $A \leftarrow A \cup \{e\}$}
    \If{$E(w_i) \subset A$}{
    $\Phi(w_i) = \textstyle \sum_{e \in E(w_i)} \tau_e$ \\
    $\Phi^{t_i}(w_i) = \textstyle  \sum_{e \in E(w_i)} \tau_e \cdot \tau^{t_i}_e$\\
    $B \leftarrow B \cup \{w_i\}$}
    }
    $B \leftarrow B \backslash \{w_j\}$
    }
    }
    \For{$w_i \in W(\calC)$}{
    \For{$e \in E(w_i) \cap E_\Lambda$}{
    $\hat{\pi}_e = \tfrac{\tau_e \cdot \tau_e^{t_i}}{\Phi^{t_i}(w_i)}$, $\hat{\pi}^t_e ={\pi}^t_e$ 
    }
    \For{$e \in E(w_i) \backslash E_\Lambda$}{
    $\hat{\pi}_e = 0$, $\hat{\pi}^t_e = 0$}
    }
    \KwRet{$\boldsymbol{\hat{\pi}}$, $\boldsymbol{\hat{\pi}}^t$ }
    \caption{The CT-CEG propagation algorithm}
\label{prop_alg_pseudo}
\end{algorithm}

\subsubsection{Incomplete Temporal Evidence}
\label{subsubsec:path_probs}

So far we considered evidence where we knew all the arrival times from the root up to the sink. However, it may happen that the evidence $E$ contains the arrival times only up to some non-sink vertex $w$. In this scenario, the t-potentials and both the emphases are as defined in Section \ref{subsec:algorithm}. The h-potential $\tau_e^{t_{w'}}$ is set to one for all vertices $w'$ for which the transition time from $w'$ is unknown. For the other vertices, define the h-potential as in Section \ref{subsec:algorithm}. Thus when the transition time for a certain vertex is not known, we revert to the standard CEG propagation algorithm for that vertex. Observe that both these types of temporal evidence automatically remove the possibility of having visited any paths that contain fewer or more vertices than we have arrival times.

In several applications, there may be directed paths of varying lengths from the root $w_0$ to some vertex $w$, denote them by $\Lambda(w_0, w) \subseteq \Lambda(w)$. Evidence $E'$ might only indicate that the unit has arrived at $w$ at some time $t = t^*$, following the convention that the arrival time at the root is $t = 0$. In such cases, often the interest is in knowing the probabilities associated with the different paths in $\Lambda(w_0, w)$. Examples of domains where this may be the case are medicine, law and criminal justice.

To obtain these path probabilities, we first construct the transporter CT-CEG $\calC^*$ and calculate the revised transition probabilities, denoted by $\hat{\pi}(.)$, only using the non-temporal evidence in $E'$ and the CEG propagation algorithm. For each path $\lambda_i \in \Lambda(w_0, w)$, let the random variable $H(\lambda_i(w_0, w))$ indicate the time it takes to get from vertex $w_0$ to $w$ in $\calC^*$, when the unit goes along path $\lambda_i$. Then $H(\lambda_i(w_0, w))$ is a convolution of the holding time densities along the edges in $\lambda_i$. The probability that the unit travelled to $w$ from $w_0$ along path $\lambda_i$ is given by
\begin{align*}
    & \pi(\lambda_i \,|\, H(\lambda_i(w_0, w)) = t^*, w, E')\\
    &= \dfrac{\pi(H(\lambda_i(w_0, w)) = t^*, w, \lambda_i, E')}{\pi(H(\lambda_i(w_0, w)) = t^*, w, E')} \\
    &= \dfrac{\pi(H(\lambda_i(w_0, w)) = t^* \,|\, w, \lambda_i, E') \hat{\pi}(\lambda_i, E' \,|\, w)}{
\sum_{\lambda_k} \pi(H(\lambda_k(w_0, w)) = t^* \,|\, w, \lambda_k, E')  \hat{\pi}(\lambda_k, E' \,|\, w)}
\end{align*}
\noindent where $\lambda_k \in \Lambda(w_0, w)$. The holding time densities are invariant under any evidence observed and the path probability $\hat{\pi}(\lambda_i, E' \,|\, w)$ is obtained as
\begin{equation*}
    \hat{\pi}(\lambda_i, E' \,|\, w) = \textstyle\prod_{e \in \lambda_i} \hat{\pi}_e.
\end{equation*}

Alternatively, revised transition probabilities could be obtained by iteratively computing the holding time density on edges upstream of $w$ given the temporal evidence in $E'$. However, this requires integrating over the possible times the unit could have arrived at the intermediate vertices which is a non-trivial task.

\section{DYNAMIC PROPAGATION}
\label{sec:propagation}

For a given CT-DCEG $\calD$, suppose the evidence $E$ pertains to a set of positions contained in passage-slices $k$ to $k+l$, $k, l \in \mathbb{N}$. Assume that $E$ is compatible with respect to the model $\calP_{k:k+l}$. Then $\calD$ can be split into three models: present, past and future. The construction and propagation scheme within each of these models is described below.

\textit{\textbf{Present model:}} The present model is given by $\calP_{k:k+l}$. Propagating the compatible evidence $E$ in this model proceeds exactly as in a static CT-CEG (see Section \ref{sec:ctceg}). Note that situations in the event tree of a CT-DCEG which have a directed path between them could be in the same position as the infinitely large subtrees rooted at them could be isomorphic. However, when a CT-DCEG is unrolled and we look at a finite number of passage-slices, they can no longer be in the same position. This is necessary for writing \ref{eq:part2} as \ref{eq:part3} in the proof in the appendix.

\textit{\textbf{Past model:}} The unrolled model $\calP_{1:k-1}$ gives the past model. Any evidence in the present model also affects the past model. However, this need not be propagated to the past model unless we need to re-estimate the past probability distributions or make inferences about the positions within the past passage-slices. 

Passage-slices from the past model can be moved to the present model in a straightforward way for re-estimating the probabilities therein. For instance, for inference on evidence $E^*$ concerning positions in passage-slice $P_i$, $i \in \mathbb{N}, i \leq k-1$, $\calP_{i:k-1}$ can be incorporated into the present model as follows. First, the revised past model is given by $\calP_{1:i-1}$. Next, the vertices and edges that are not visited with probability one, conditioned on evidence $E$ and $E^*$, are deleted from $\calP_{i:k-1}$. Denote this by $\calP^*_{i:k-1}$. For an edge $e \in \epsilon_{k-1}$, if $e \in E(\calP^*_{i:k-1})$, then connect it to the relevant vertex in the present model. This gives us the revised present model. Propagation continues backwards from the root vertices of the original present model to the root vertices of the revised present model. 

\textit{\textbf{Future model:}} The graph of a finite CT-DCEG as it applies to passage-slices $P_t$, $t > k+l$ is first adapted to delete all the edges and vertices that will not be visited in future passage-slices with probability one given the evidence $E$. Call this $\calD^*$. The conditional transition probabilities at each position $w \in W(\calD^*)$ are revised as
\begin{align*}
    \hat{\pi}_e(w'\,|\,w) = \dfrac{\pi_e(w'\,|\,w)}{\sum_{e' \in E(w)}\pi_{e'}(w'\,|\,w)}.
\end{align*}
\noindent Recall that the holding time distributions defined along the extant edges in $E(\calD^*)$ are invariant under observing evidence and can be imported directly from $\calD$. The adapted CT-DCEG can now be represented by the state-transition diagram of a, possibly condensed, SMP (see Section \ref{subsec:semi-markov}). Forecasts concerning probabilities of future events are calculated using the transition matrix of its embedded Markov chain. Additionally, all inferences that can be typically made from a semi-Markov process or a CT-DCEG can still be made in the standard way \citep{shenvi2019bayesian}. 

The above scheme, although simple, is capable of making a wide range of inferences. While it is analogous to the dynamic inference scheme for DBNs in \citet{kjaerulff1992computational}, movement between the three models is much easier for the CT-DCEG. This is because we do not need to reconstruct a junction tree and propagation is carried out directly on the vertices and edges of the adapted graphs of the concerned model. More importantly, the complexity of propagation in the past and present models is \textit{linear} in the number of vertices they contain, whereas it is exponential in the maximal clique size in a junction tree. Another key advantage of the CT-DCEG is that additional intrinsic events observed always lead to a simplification of the graph and thereby to efficiency gains.

%In the CT-DCEG, additional compatible evidence (see Section \ref{subsec:compatibility}) does not lead to any additions to the original graph topology. In fact, in most cases, it vastly simplifies the graph as the probability of not visiting certain vertices and edges becomes one. We believe this is a key advantage of using tree-based models for modelling dynamic processes like these.

\section{APPLICATIONS}
\label{sec:applications}

\subsection{A SIMPLE APPLICATION}

We now revisit Example 2. Suppose that for an individual who has had the infection twice in the past, we observe that they had the infection again, were treated for it and recovered from it for the third time. Suppose we also observe that the individual had the following transition times $t_1 = 2.5, t_2 = 6.5, t_3 = 11$ recorded as the number of days since they recovered last time $t_0 = 0$. For this evidence, the present model is simply given by Figure \ref{fig:transporter} and propagation in this model requires only 32 operations: 8 t-potentials, 8 h-potentials, 5 t-emphases, 5 h-emphases and 6 revised edge probabilities.

\begin{table}[ht]
\caption{Conditional transition probabilities and holding time distributions for Example 2 [$H_{i,j} \triangleq H(e(w_i, w_j))$; M: multinomial; E: exponential; N: normal; W: weibull]}
\label{table:distributions_example}
\begin{center}
\begin{tabular}{l|l}
$X(w_0) \sim M(0.4,0.3,0.3) $     & $H_{1,3} \sim N(7,1)$\\
$X(w_1) \sim M(0.45,0.55) $     & $H_{1,4} \sim N(5,2)$ \\
$X(w_2) \sim M(0.9,0.1) $     & $H_{4,0}\sim W(2.8,30)$\\
$X(w_3) \sim M(0.73,0.27) $     & $H_{4,\infty} \sim W(0.8,1.5)$ \\
$X(w_4) \sim M(0.8,0.2)$      & $H_{3,0} \sim W(1.8,24)$ \\
\{Strain 1\} \, $H_{0,1} \sim E(2)$    & $H_{3,\infty} \sim W(0.88,2)$ \\
\{Strain 2\} \, $H_{0,1} \sim E(2.8)$  & $H_{2,0}\sim W(1.3,12)$\\
$H_{0,2} \sim E(3.5)$  & $H_{2,\infty} \sim W(0.7,1.8)$ \\
\end{tabular}
\end{center}
\end{table}

\begin{figure}[ht]
%\vspace{1in}
\centering
\includegraphics[trim = 7cm 19cm 8cm 0cm, scale = 0.25 ]{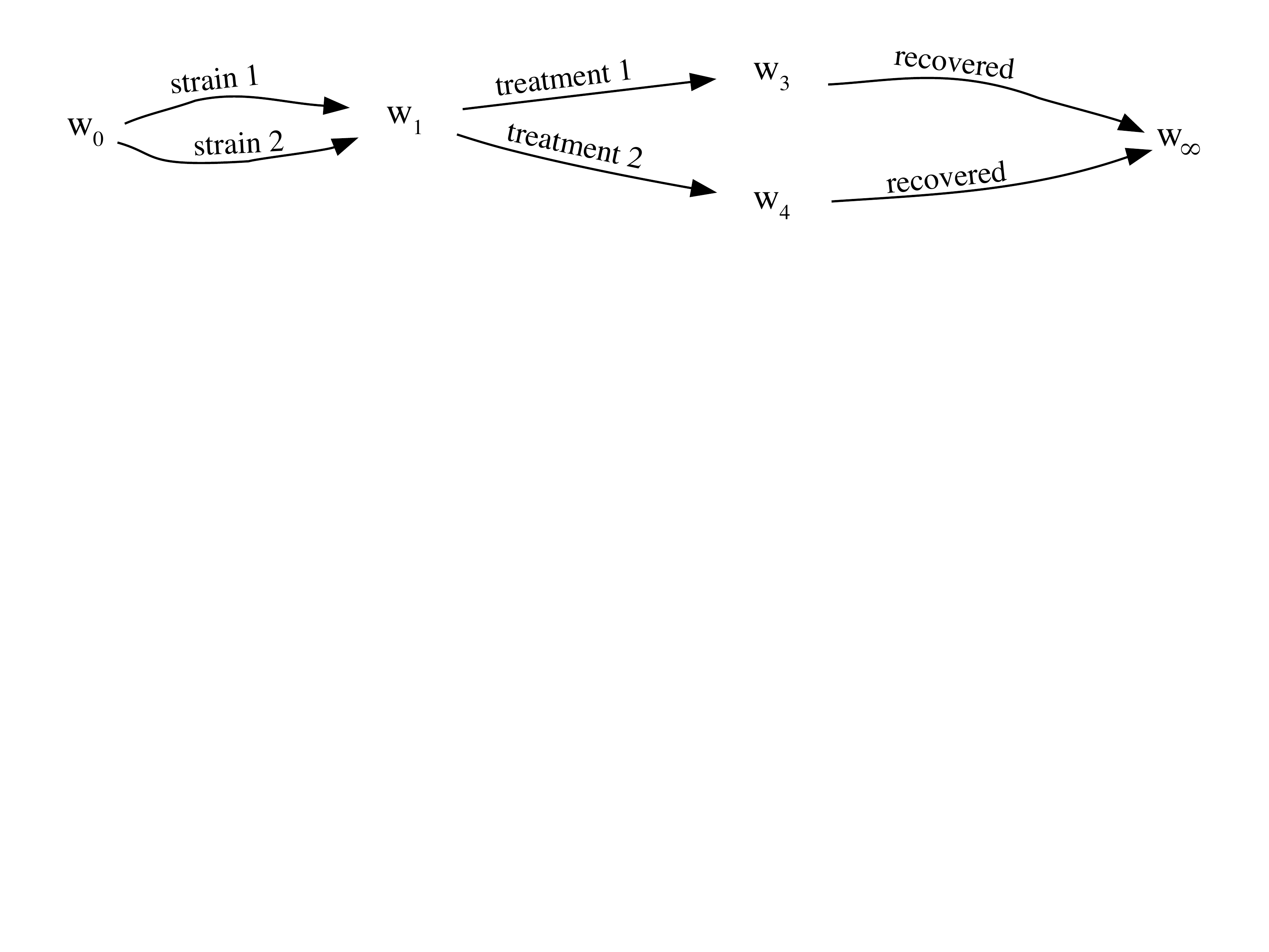}
\caption{The present model}
\label{fig:transporter}
\end{figure}
%Add the adjusted info directly into the image here.

\begin{figure}[ht]
\vspace{1in}
\centering
\includegraphics[trim = 7cm 12cm 8cm 15cm, scale = 0.16 ]{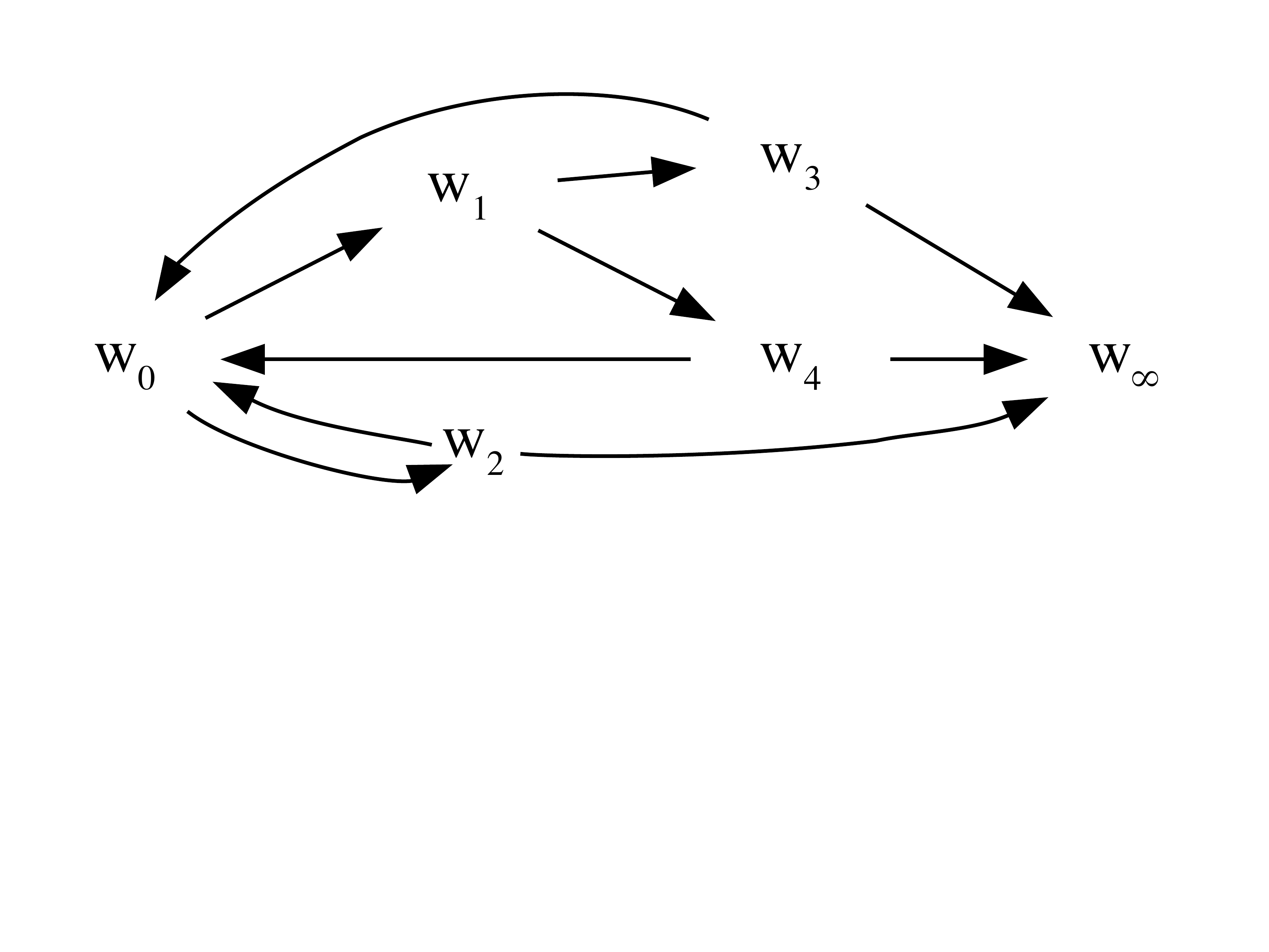}
\caption{The future model}
\label{fig:smp}
\end{figure}

The future model is represented by an SMP whose state-transition diagram is given in Figure \ref{fig:smp}. The two edges from $w_0$ to $w_1$ in Figure \ref{fig:transporter} are replaced by a single edge in the SMP. As reinfection from any of the strains is possible, the CT-DCEG for the future model is identical to the original CT-DCEG in Figure \ref{fig:ct-dceg}. Observe that $w_\infty$ acts as an absorbing state in the SMP. This example is explored in greater detail in the supplementary material.

\begin{figure}[ht]
%\vspace{1in}
\centering
\includegraphics[trim = 2cm 13.5cm 8cm 1cm, scale = 0.28 ]{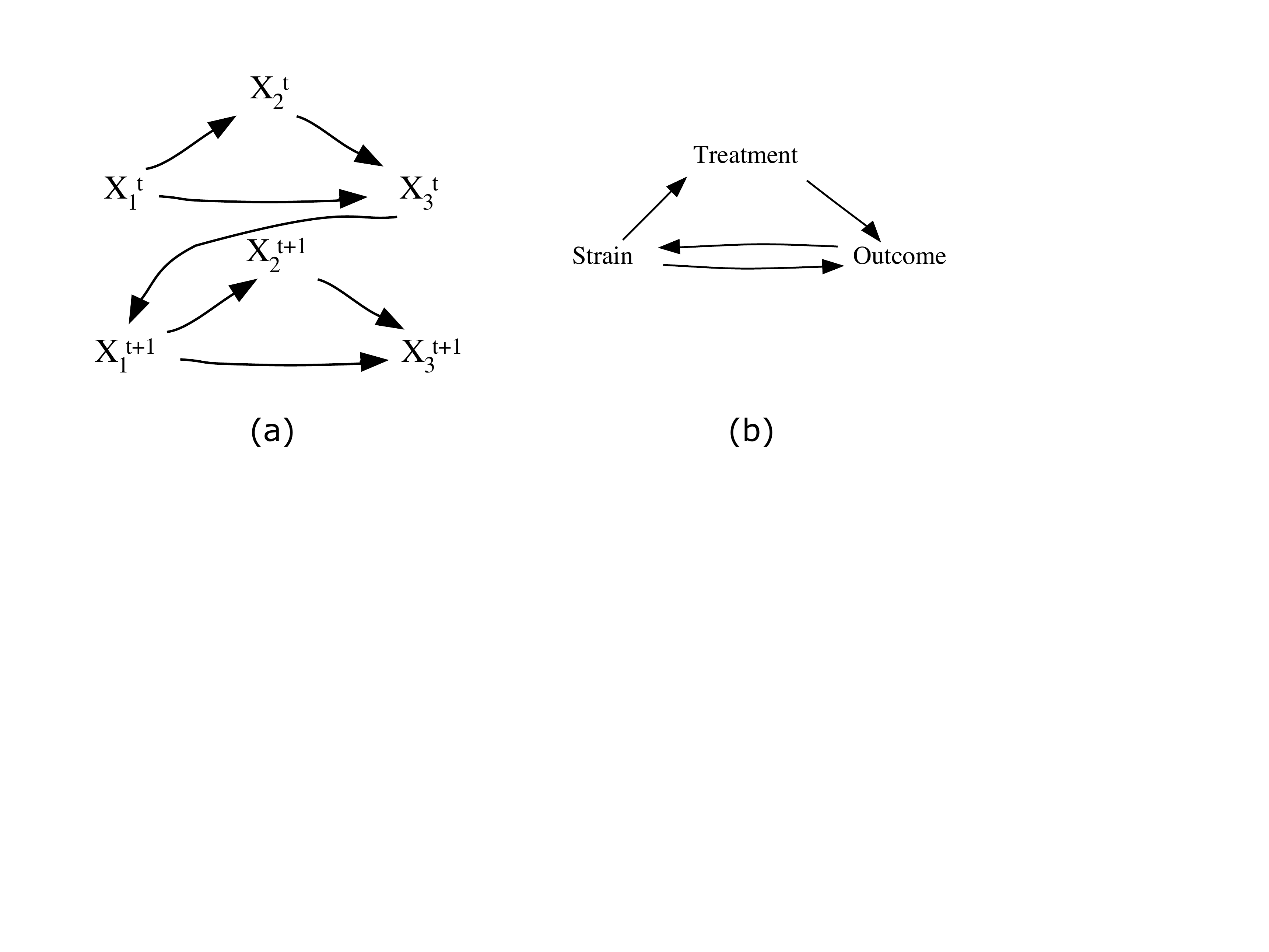}
\caption{Example 2 as (a) a DBN; (b) a CTBN}
\label{fig:ctbn_dbn}
\end{figure}

It is instructive to compare this problem representation with alternative dynamic models: the DBN and the CTBN. A DBN for Example 2 could be represented by the two time-slice DBN in Figure \ref{fig:ctbn_dbn}(a) where the variables $X_1$, $X_2$ and $X_3$ represent the strain of the infection, the treatment type and outcome respectively. Given the significant asymmetries in the example, this DBN is an \textit{approximation} and hides away structural zeros within its CPTs. It also does not graphically represent the lack of treatment options for strain 3 of the infection. 
%Structural zeros are a logical impossibility rather a sampling limitation.
%Even for the simple evidence above, we would require  (insert number of junction tree calculations).

Figure \ref{fig:ctbn_dbn}(b) shows a CTBN for Example 2. Due to structural zeroes, some of the conditional intensity matrices are null matrices (e.g. for treatment given the third strain of infection). Additionally, as seen from Table \ref{table:distributions_example}, the process contains non-exponential holding times which CTBNs were not designed to represent: CTBN propagation algorithms rely on exploiting the exponential nature of the holding times. 

\subsection{A MIXED CEG APPLICATION}
\label{subsubsec:mixed}

For certain types of transitions, it is not natural to define a holding time at that vertex. For instance, a vertex categorising an individual into different risk categories may not be meaningfully associated with a holding time. Call a CEG a \textit{mixed CEG} if its vertex set can be partitioned into two mutually exclusive subsets $V_1$ and $V_2$ such that transitions from vertices in $V_1$ are associated with holding times and those from vertices in $V_2$ are not. However, using a mixed CEG is a modelling choice. The modeller may choose to associate such a vertex with, for example, the time it takes a health practitioner to make the categorisation. Additionally, $V_1$ may contain vertices with holding time distributions in discrete as well as continuous time domains. In our experience such mixed systems are more common in the real world than a homogeneously defined one. In fact, the public health applications considered in \citet{shenvi2019bayesian} are mixed DCEGs although they were not recognised as such. For illustrative purposes, we consider one of these applications to emphasise the usefulness of mixed (D)CEGs.

\begin{figure}[ht]
%\vspace{1in}
\centering
\includegraphics[trim = 2.5cm 8cm 2.5cm 2cm, scale = 0.27 ]{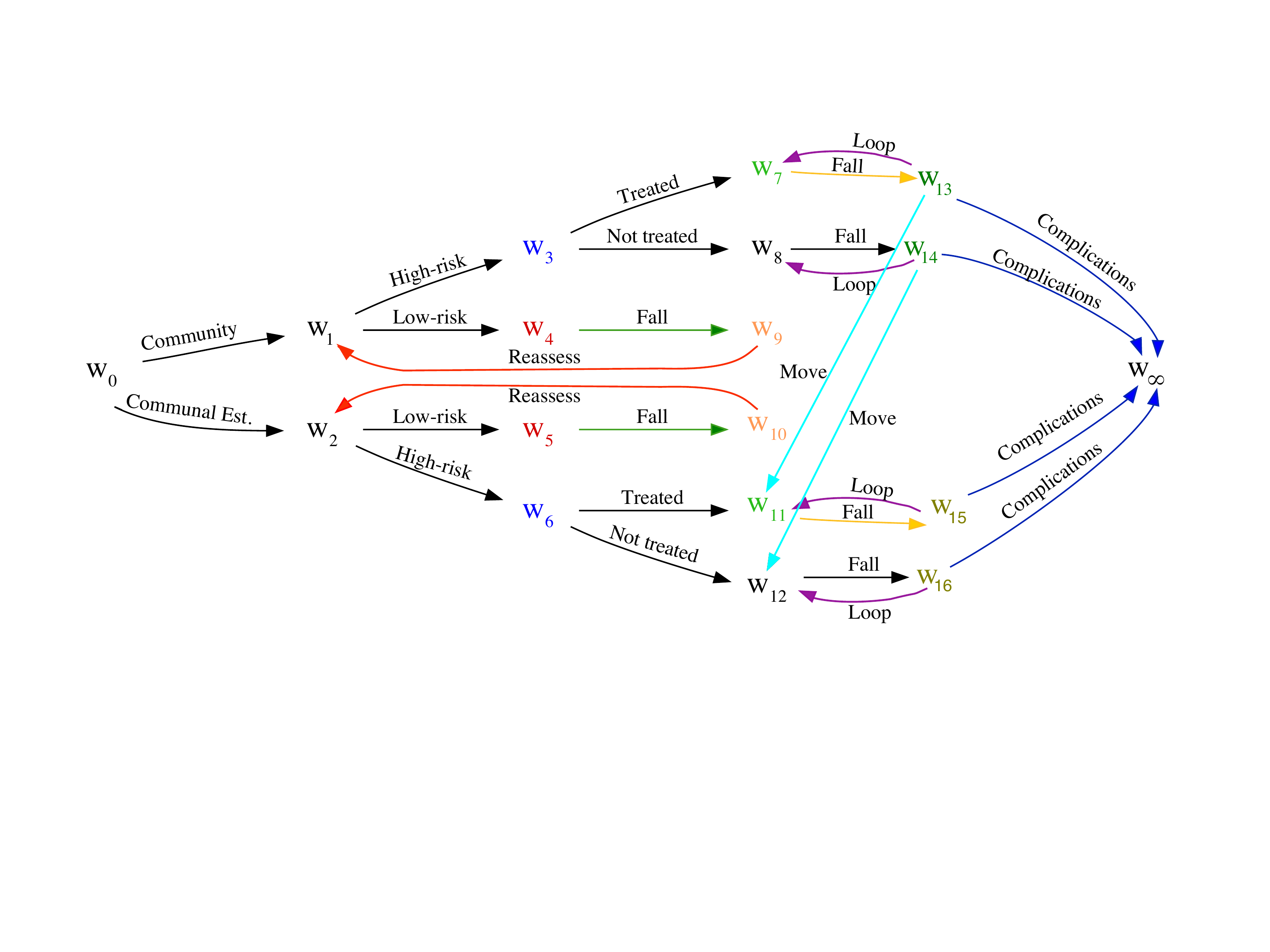}
\caption{A mixed DCEG}
\label{fig:falls_rdceg}
\end{figure}

\textbf{Example 3:} Consider the mixed DCEG in Figure \ref{fig:falls_rdceg} based on a real-world falls intervention \citep{eldridge2005modelling}. The transitions from $w_i$, $i = 0, 1, 2, 3, 6$ describe categorical events that are not naturally associated with holding times. However, all transitions from the remaining vertices are best described in conjunction with \textit{how long} it took for such transitions to occur and such descriptions are of clinical importance. 

For propagation in a mixed CEG, the h-potentials for edges emanating from vertices in $V_2$ are set to be one. The remaining potentials and emphasis are defined as in Section \ref{subsec:algorithm}. Thus our methodology can be adapted in a straightforward way to the wide range of applications for which mixed CEGs are appropriate.

\section{DISCUSSION}
\label{sec:discussion}

We presented an evidence propagation scheme for the CT-DCEG - a temporal, dynamic event-based graphical model - within a statistically grounded framework. This scheme is also applicable to all other DCEGs. We also filled the technical gaps in the literature for such a scheme to work by providing a propagation algorithm for a static CT-CEG, and briefly explored the novel class of mixed CEGs.

We have demonstrated how the CT-DCEG gives a better representation of a process when it contains significant asymmetries and when the evolution of the process can be described under some total ordering of the events. For dynamic processes with fewer asymmetries, where co-evolution of components is important and where holding times are known to be exponential, the CTBN should be preferred. On the other hand, the DBN is the ideal and most well-developed modelling tool available when the process meets all the conditions for the CTBN and all the components can be assumed to evolve at the same discrete time granularity.

Finally, in this paper we only looked at certain types of point temporal evidence. Methods to incorporate interval temporal evidence in the CT-(D)CEG are yet to be devised. Additionally, certain types of point and interval temporal evidence can lead to confounding by making all the past subprocesses highly dependent on each other. One example of this is addressed in a very simple way in Section \ref{subsubsec:path_probs}. In case of such evidence we suggest (a) deferral of its inclusion until further evidence is obtained, or (b) use of approximate inference schemes. The latter remains an open problem for this class of models. 

%talk about the rate at which an individual moves through the system.

\section*{APPENDIX}

The CT-CEG propagation algorithm states that for $\hat{\pi}_e(w' \,|\, w) \triangleq \pi( \Lambda(e(w, w')) \,|\, E, \Lambda(w))$ we have
\begin{align*}
        \hat{\pi}_e(w' \,|\, w) =
        \begin{cases}
        \dfrac{\tau_e(w' \,|\, w) \, \tau_e^{t_w}(w' \,|\, w)}{\Phi^{t_w}(w)}, &\\
        \quad \quad \quad \quad \textmd{if}\, e(w, w') \in E(\calC^*)\\
        0, & \\
        \quad \quad \quad \quad \textmd{if}\, e(w, w') \notin E(\calC^*).
        \end{cases}
    \end{align*}
where the potentials and emphases are as defined in Section \ref{subsec:algorithm} and $E$ is some compatible evidence.

\textbf{Proof:}\\
Given  that a unit is in some position $w$, the probability of transitioning along an edge emanating from $w$ is independent of the holding time at $w$. This gives us
\begin{align*}
    \hat{\pi}_e(w' \,|\, w) &= \pi( \Lambda(e(w, w')) \,|\, E, H_w = t_w, \Lambda(w)) \\
    &= \dfrac{\pi( \Lambda(e(w, w')), E, H_w = t_w, \Lambda(w))}{\pi(E, H_w = t_w, \Lambda(w))} \stepcounter{equation}\tag{\theequation}\label{eq:part1}
\end{align*}

Let $\calT$ denote the event tree underlying the CT-CEG $C$ and $\calT^*$ - a subtree of $\calT$ - denote the event tree underlying the transporter CT-CEG $C^*$. By the definition of a position, each $w \in W(\calC)$ corresponds to a set of vertices $\{v_i\}$ in $\calT$. Then, for $w$, we can split this into two mutually exclusive subsets: $\{v_i\}_{i \in I}$ representing the vertices of $\{v_i\}$ in $\calT^*$ and $\{v_i\}_{i \in J}$ representing the vertices of $\{v_i\}$ not in $\calT^*$. Additionally, the paths in $\Lambda(w)$ and $\Lambda(e(w, w'))$ in $\calC$ are a union of the paths in $\Lambda(v_i)$ and $\Lambda(e(v_i, v_i'))$ in $\calT$ respectively for $i \in I \cup J$. Every $v_j \in \{v_i\}_{i \in I \cup J}$ represents $w \in W(\calC)$. Thus, for some $w'$ represented by $v_j'$, we have
\begin{align*}
    \pi_e(v_j' \,|\, v_j) &= \pi_e(w' \,|\, w),\\
    \pi_e^{t_{v_j}}(v_j' \,|\, v_j) &= \pi_e^{t_w}(w' \,|\, w), \quad \textmd{for}\,\, t_{v_j} = t_w.
\end{align*}

This allows us to write (\ref{eq:part1}), for $t_{v_i} = t_w$, as
\begin{align}
    & \dfrac{\pi(E, \cup_{i \in I \cup J} [\Lambda(e(v_i, v_i')), H_{v_i} = t_{v_i}, \Lambda(v_i)])}{\pi(E, \cup_{i \in I \cup J} [H_{v_i} = t_{v_i}, \Lambda(v_i)])},
\label{eq:part2}
\end{align}
\noindent to be evaluated on the tree $\calT$. There is no directed path from $v_j$ and $v_k$ for $v_j, v_k \in \{v_i\}_{i \in I \cup J}, j \neq k$ as their subtrees are isomorphic in $\calT$. Hence we have that $\Lambda(v_j) \cap \Lambda(e(v_k, v_k')) = \emptyset$. So we can write (\ref{eq:part2}) as
\begin{align}
    & \dfrac{ \sum_{i \in I \cup J} \pi(E, \Lambda(e(v_i, v_i')), H_{v_i} = t_{v_i}, \Lambda(v_i))}{ \sum_{i \in I \cup J} \pi(E, H_{v_i} = t_{v_i}, \Lambda(v_i))}.
\label{eq:part3}
\end{align}

For $v_j \in \{v_i\}_{i \in J}$ we have that $\calT^*_\Lambda \cap \Lambda(v_j) = \emptyset$. Also, notice that writing $\pi(.)$ as a probability over paths of $v_j$ is equivalent to writing it over paths of $v_k$, for $v_j, v_k \in \{v_i\}_{i \in I}$. So we can write (\ref{eq:part3}) as
\begin{align*}
    & \dfrac{ \sum_{i \in I} \pi(E, \Lambda(e(v_i, v_i')), H_{v_i} = t_{v_i}, \Lambda(v_i))}{ \sum_{i \in I} \pi(E, H_{v_i} = t_{v_i}, \Lambda(v_i))}\\
    &= \Bigg \{\dfrac{\pi(H_{v_j} = t_{v_j} \,|\, E, \Lambda(e(v_j, v_j')), \Lambda(v_j))}{\pi(E, H_{v_j} = t_{v_j} \,|\, \Lambda(v_j))} \times \\
    & \dfrac{\pi(E, \Lambda(e(v_j, v_j')) \,|\, \Lambda(v_j)) \sum_{i \in I}\pi(\Lambda(v_i))}{\sum_{i \in I}\pi(\Lambda(v_i))} \Bigg \}\\
    &= \Bigg \{\dfrac{\pi(H_{v_j} = t_{v_j} \,|\, E, \Lambda(e(v_j, v_j')), \Lambda(v_j))}{\pi(E, H_{v_j} = t_{v_j} \,|\, \Lambda(v_j))} \times \\
    & \pi(E, \Lambda(e(v_j, v_j')) \,|\, \Lambda(v_j)) \Bigg \}
\end{align*}
\noindent for any $v_j \in \{v_i\}_{i \in I}$.
 
The proofs for $\tau_e(w'\,|\,w) = \pi(E, \Lambda(e(v_j, v_j')) \,|\, \Lambda(v_j))$ where $e(w, w') \in C^*$, $v_j \in \{v_i\}_{i\in I}$, and $\Phi(w) = \pi(E\,|\, \Lambda(v_j))$ follow exactly as given in \citet{thwaites2008propagation}. Additionally, we have for $t_{v_j} = t_w$,
 \begin{align*}
     \tau_e^{t_w}(w'\,|\,w) &= \pi_e^{t_w}(w' \,|\,w) \\
     &= \pi(H_{v_j} = t_{v_j} \,|\, \Lambda(e(v_j, v_j')), \Lambda(v_j)) \\
     &= \pi(H_{v_j} = t_{v_j} \,|\, E, \Lambda(e(v_j, v_j')), \Lambda(v_j)),
 \end{align*}
 \noindent by definition of $\tau_e^{t_w}(w'\,|\,w)$ and by the invariance of the holding time density given any compatible evidence $E$. We use induction to prove that $\Phi^{t_w}(w) = \pi(E, H_{v_j} =t_{v_j} \,|\, \Lambda(v_j))$ where $v_j \in \{v_i\}_{i \in I}, t_{v_j} = t_w$, $\forall w \in W$. 
 
 \textbf{Step 1:} Consider the positions $w \in W(-1)$. We have
\begin{align*}
    \Phi^{t_w}(w) &= \sum_{e \in E(w)} \tau_e(w_\infty| w) \, \tau_e^{t_w}(w_\infty|w) \\
    &= \sum_{e \in E(w)}  \pi_e(w_\infty| w) \, \pi_e^{t_w}(w_\infty|w) \\
    &= \sum_{e \in E(v_j)}  \pi_e(v_{\textmd{leaf}}| v_j) \, \pi_e^{t_{v_j}}(v_{\textmd{leaf}}|v_j) \\
    &= \pi(E \,|\,\Lambda(v_j)) \,\pi(H_{v_j} = t_{v_j} \,|\, \Lambda(v_j), E)\\
    &= \pi(E, H_{v_j} = t_{v_j} \,|\, \Lambda(v_j)),
\end{align*}
\noindent for any $v_j \in \{v_i\}_{i \in I}$ and $t_{v_j} = t_w$.

\textbf{Step 2:} Now consider any $w \in W$ such that all the vertices $\{w'\}$ into which $E(w)$ terminate have $\Phi^{t_{w'}}(w') = \pi(E, H_{v_j'} = t_{v_j'} \,|\, \Lambda(v_j'))$. Then we have
\begin{align*}
    \Phi^{t_w}(w) &= \sum_{e\in E(w)} \tau_e(w'|w) \, \tau_e^{t_w}(w'|w) \\
    &= \sum_{e\in E(w)} \pi_e(w'|w) \, \Phi(w') \, \pi_e^{t_w}(w'|w) \\
    &= \sum_{e \in E(v_j)} \pi_e(v_j'|v_j) \, \pi(E|\Lambda(v_j')) \, \pi_e^{t_{v_j}}(v_j'|v_j). 
\end{align*}
However, in a tree, we have that $\Lambda(v_i') = \Lambda(e(v_i, v_i')) \subset \Lambda(v_i)$. So we can write $\Phi^{t_w}(w)$ as 
\begin{align*}
    & \sum_{e \in E(v_j)} \Big \{\pi(\Lambda(e(v_j, v_j')), \Lambda(v_j')|\Lambda(v_j)) \times \\
    & \pi(E | \Lambda(v_j'), \Lambda(e(v_j, v_j')), \Lambda(v_j)) \times \pi_e^{t_{v_j}}(v_j'|v_j) \Big \}\\
    &= \sum_{e \in E(v_j)} \pi(E, \Lambda(v_j'), \Lambda(e(v_j, v_j')) |\Lambda(v_j))\, \pi_e^{t_{v_j}}(v_j'|v_j)  \\
    &= \sum_{e \in E(v_j)} \pi(E, \Lambda(e(v_j, v_j')) \,|\, \Lambda(v_j)) \, \pi_e^{t_{v_j}}(v_j'|v_j)  \\
    &= \pi(E, \Lambda(v_j)\,|\, \Lambda(v_j)) \,\, \pi(H_{v_j} = t_{v_j} \,|\, \Lambda(v_j), E) \\
    &= \pi(E, H_{v_j} = t_{v_j} \,|\, \Lambda(v_j)).
\end{align*}

This completes the proof. 

\section*{SUPPLEMENTARY MATERIAL}

\renewcommand{\thesection}{\Alph{section}}
\setcounter{section}{0}
\section{Propagation in the present model}

In this section, we continue our analysis of Example 2 given the evidence in Section 5.1. The potentials and emphases for the present model are given in Tables \ref{table:potentials} and \ref{table:emphases} respectively. Here $e^i(w_0, w_1)$ refers to the edge associated with strain $i$ of the infection, $i = 1, 2$.

\begin{table}[ht]
\caption{Potentials in Example 2}
\label{table:potentials}
\begin{center}
\begin{tabular}{c|c|c}
\multicolumn{1}{c|}{\bf Edge}  &\multicolumn{1}{c|}{\bf t-potential} & \multicolumn{1}{c}{\bf h-potential}  \\
\multicolumn{1}{c|}{$e(w, w')$} & \multicolumn{1}{c|}{$\tau_e(w'\,|\,w)$} & \multicolumn{1}{c}{$\tau_e^t(w'\,|\,w)$} \\
\hline
$e(w_3, w_\infty)$ & $0.73$ & $t = 4.5$, \,\, $0.17826$  \\
$e(w_4, w_\infty)$ & $0.80$ & $t = 4.5$, \,\, $0.91921$  \\
$e(w_1, w_3)$ & $0.3285$ & $t = 4$, \,\, $0.00443$  \\
$e(w_1, w_4)$ & $0.44$ & $t = 4$, \,\, $0.17603$  \\
$e^1(w_0, w_1)$   & $0.3074$ & $t = 2.5$, \,\, $0.01348$  \\
$e^2(w_0, w_1)$ & $0.23055$ & $t = 2.5$, \,\,$0.00255$  \\
\end{tabular}
\end{center}
\end{table}

\begin{table}[ht]
\caption{Emphases in Example 2}
\label{table:emphases}
\begin{center}
\begin{tabular}{c|c|c}
\multicolumn{1}{c|}{\bf Vertex}  &\multicolumn{1}{c|}{\bf t-emphasis} & \multicolumn{1}{c}{\bf h-emphasis}  \\
\multicolumn{1}{c|}{$w$} & \multicolumn{1}{c|}{$\Phi(w)$} & \multicolumn{1}{c}{$\Phi^t(w)$} \\
\hline
$w_3$ & $0.73$ & $t = 4.5$, \,\, $0.13013$  \\
$w_4$ & $0.80$ & $t = 4.5$, \,\, $0.73537$  \\
$w_1$ & $0.7685$ & $t = 4$, \,\, $0.07891$  \\
$w_0$ & $0.5380$ & $t = 2.5$, \,\, $0.00473$  \\
\end{tabular}
\end{center}
\end{table}

The revised transition probabilities are shown in Figure \ref{fig:transporter}. Let paths $\lambda_i$ $ i = 1, 2, 3, 4$ be given by the sequences of edges

\begin{align*}
    \lambda_1 &= (e^1(w_0, w_1), e(w_1, w_3), e(w_3, w_\infty)) \\
    \lambda_2 &= (e^1(w_0, w_1), e(w_1, w_4), e(w_4, w_\infty)) \\
    \lambda_3 &= (e^2(w_0, w_1), e(w_1, w_3), e(w_3, w_\infty)) \\
    \lambda_4 &= (e^2(w_0, w_1), e(w_1, w_4), e(w_4, w_\infty)). 
\end{align*}

\begin{figure}[ht]
\centering
\includegraphics[trim = 7cm 19cm 8cm 0cm, scale = 0.25 ]{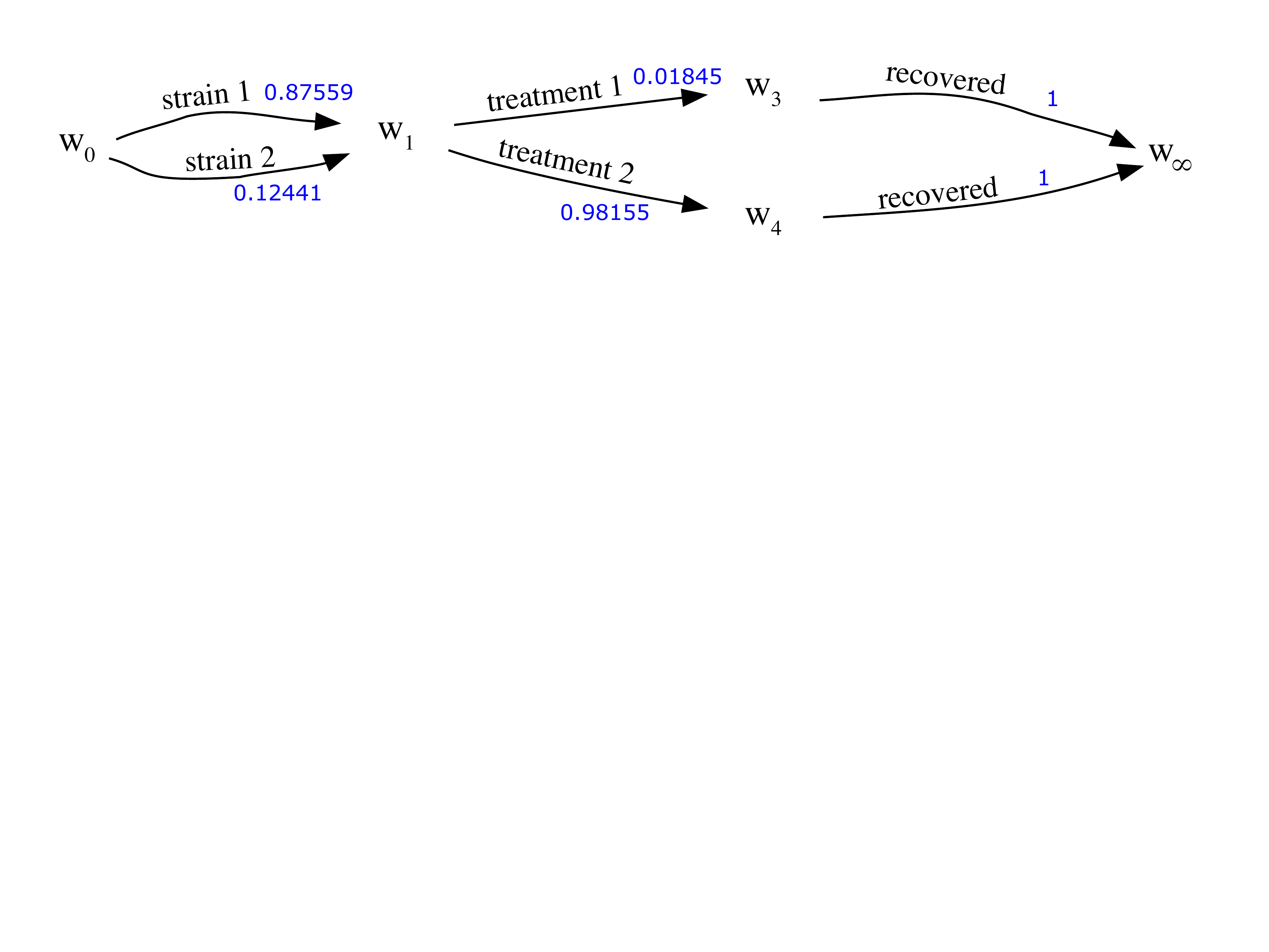}
\caption{Present model with the revised transition probabilities}
\label{fig:transporter}
\end{figure}

The probabilities $\pi(\lambda_i)$ that the individual went along path $\lambda_i$,  $i = 1, 2, 3, 4$ are given in Table \ref{table:path_probs}. For comparison, we also show what the path probabilities would have been if we had not corrected them for the temporal evidence and had only used the non-temporal parts of the evidence. These are given in the table under the column $\pi^*(\lambda_i)$. This clearly shows how knowing the timing of the transitions conveys essential information about the evolution of the process. 

\begin{table}[ht]
\caption{Path probabilities in Example 2}
\label{table:path_probs}
\begin{center}
\begin{tabular}{c|c|c}
\multicolumn{1}{c|}{\bf Path $\lambda_i$}  &\multicolumn{1}{c|}{\bf $\pi(\lambda_i)$} & \multicolumn{1}{c}{\bf $\pi^*(\lambda_i)$} \\
\hline
$\lambda_1$ & $0.01615$ & $0.24426$  \\
$\lambda_2$  & $0.85944$ &  $0.32717$  \\
$\lambda_3$  & $0.00230$ & $0.18320$  \\
$\lambda_4$  & $0.12211$ & $0.24537$  \\
\end{tabular}
\end{center}
\end{table}

\newpage
\subsubsection*{References}
\renewcommand{\refname}{\vspace{-4ex}}
\bibliography{biblio}
\bibliographystyle{apalike}

\end{document}